\documentclass[11pt]{article}
\usepackage{hyperref}
\usepackage[a4paper, lmargin=1.0in, rmargin=1.0in, tmargin=1.0in, bmargin=1.0in]{geometry}
\begin{document}


\title{Design of a P System based Artificial Graph Chemistry}
\author{{\bf Janardan Misra}~\thanks{Work done when author was in NUS (2002-2005).} \\ HTS (Honeywell Technology Solutions) Research Lab\\ 151/1 Doraisanipalya, Bannerghatta Road,\\ Bangalore 560 076, India\\ {\sf Email: janardan.misra@honeywell.com} }

\date{}
\maketitle
\tableofcontents

\begin{abstract}
Artificial Chemistries (ACs) are symbolic chemical metaphors for
the exploration of Artificial Life, with specific focus on the origin of
life. In this work we define a P system based artificial graph chemistry to
understand the principles leading to the evolution of life-like
structures in an AC set up and to develop a unified framework to
characterize and classify symbolic artificial chemistries by
devising appropriate formalism to capture semantic and
organizational information. An extension of P system is considered by associating
probabilities with the rules providing the topological framework for the evolution of a 
labeled undirected graph based molecular reaction semantics.
\end{abstract}

\section{Basic Framework of Artificial Chemistries}

Aim of this section is to present a brief introduction to
artificial chemistries. We will start with a discussion on the epistemological
foundations of the area and will illustrate further details using examples
relevant to this proposal. The examples are followed by discussions
to motivate the main theme of the proposal which is elaborated in
coming sections.

\subsection{Introduction}

It is a long held topic of scientific debate whether there are any
biological principles of life and other complex biological
phenomena, which are not directly reducible to physical and
chemical laws. Living beings, however small and consisting of the
same molecular components as non living things, nonetheless
exhibit qualitatively different characteristics. This may be in
part due to the complex organizational structure which
distinguishes them or it could be because of their quantitatively
complex structure which gives rise to difficulty in analyzing
properties using currently available tools.

The direct ways to understand this complex biological phenomena
are usually difficult and error prone because living structures
are by default complex and hard to manipulate. Even cellular level
experiments are difficult to carry out and their simulations are
quite cumbersome.

\emph{Artificial life} (AL) is a tool to study principles
explaining this complex phenomena of life without directly getting
involved with the real biological systems. The fundamental
assumption here is that principles of life are independent of the
medium and carbon based life on earth is just one example of the
possible forms of life. This means even artificial environments
like digital media can also exhibit life-like behavior under
certain conditions. This way AL complements the main stream
biological studies by synthesizing life-like systems using digital
media. There are several such examples where these artificial life
forms exhibit properties remarkably close to higher forms of life,
e.g., Tierra \cite{tierra}, Avida \cite{1}.

Living phenomena has several aspects to study, one such is the
origin of life or \emph{biogenesis}. Here the problem is to
understand how first primitive form of life such as metabolism and
self replicating structures could have come into existence
starting from non living chemical compounds. \emph{Artificial
chemistries} (AC) are the primary tools in AL studies aimed at
understanding this origin of life and other complex emergent
phenomena. ACs follow chemical metaphor. Like real chemical
reactions between molecules, which give rise to new molecules, ACs
as well define abstract molecules and reactions and study what
emerges during the course of reactions.

An AC has three main components, a set of objects or
\emph{molecules}, a set of \emph{reaction rules} or collision
rules, and a definition of \emph{population dynamics}.

Objects can be abstract symbols, numbers, lambda expressions,
binary strings, character sequences, abstract data
structures etc. Reaction rules might be string matching,
string concatenation, reduction rules, abstract finite state
machines, Turing machines, matrix multiplication, simple
arithmetic operation, cellular automata, boolean networks etc.
Dynamics can be specified in terms of ordinary differential
equation, difference equation, meta dynamics, explicit collision
simulation, well stirred reactor, self organizing topology, etc.

A survey on various ACs is given in \cite{12}, which also has some
broad classification of ACs based upon the kind of molecular
abstractions (explicit or implicit), type of reaction rules
(constructive or non constructive), and population dynamics.

To illustrate, we take examples from two kinds of ACs. One where
no spatial structures are considered, that is, all molecules
evolve as a whole in a reactor tube and all molecules can interact
with each other according to the collision rules. The examples of
AlChemy (Section~\ref{alchemy}) and CHAM/ARMS (Section~\ref{cham}) are of this
type. Second kind of AC systems impose some sort of spatial
structures on the molecules thus limiting the possible reactions
between molecules to their ``neighborhood'' only. Planar graph
(Section~\ref{pgraph}) based AC is of this type.

It seems, during the pre-biotic evolution of life, spatial
structures (e.g., membranes etc) emerged starting from the open
reactor type system without any spatiality. This spetial structure
based classification is one of the main motivations for P system
based AC definition, we propose in the next section.

\subsection{Examples}

Next we illustrate the common design of ACs using examples. Each example is followed by a discussion on the relative strengths and limitations of it w.r.t. real chemistry. 

\subsubsection{Algorithmic Chemistry - AlChemy}~\label{alchemy}
We consider $\lambda$ expression based AC proposed in \cite{3,4}
called AlChemy.\\\\
\textbf{Molecules - $\lambda$ Terms:} The object space consists of abstract lambda expressions
 (also called \emph{terms}). These terms are generated as follows: There is an infinite supply of variable names $V$ = $\{x, y, z,\ldots\}$. Other than $V$, the alphabet consists of a lambda symbol '$\lambda$', dot '$.$', and encapsulating brackets '$($', '$)$'.

  The set of terms, $\Lambda$, is defined inductively:

(1) $x \in V \Rightarrow x \in \Lambda$

(2) $x \in V; M \in \Lambda \Rightarrow \lambda x.M \in \Lambda$
(abstraction)

(3) $M \in \Lambda; N \in \Lambda \Rightarrow (M)N \in \Lambda$
(application)

A variable $x$ is said to be bound if it occurs inside a sub-term
with the form $\lambda x.P$, otherwise it is free. The set of free
variables in an expression $P$ is denoted by $f(P)$.

\emph{Syntactical Transformation: }The schemes of transformation
are oriented rewrite rules. Structures on the left ­ hand side are
replaced by structures on the right ­ hand side. More precisely,

\emph{Substitution}

(4) $(\lambda x.x)Q \rightarrow Q$

(5) $(\lambda x.E)Q \rightarrow E$; if $x \not\in f(E)$

(6) $(\lambda x.\lambda y.E)Q \rightarrow \lambda y.(\lambda
x.E)Q$; if $x \neq y$ and $(x \not\in f(E) \vee y \not\in f(Q))$

(7) $(\lambda x.(E_1 )E_2 )Q \rightarrow ((\lambda x.E_1
)Q)(\lambda x.E_2 )Q$

\emph{Renaming}

(8) $\lambda x.E \rightarrow \lambda z.(\lambda x.E)z; z \not\in
f(E)$\\\\
\textbf{Reaction Rules - Function Composition and Normal Form
Reduction:} The reaction rules in Alchemy consist of application of one lambda
term over the other, which is then reduced to a normal form. The
choice of lambda calculus allows the abstract formulation of
chemical substitution during chemical reactions. Normalization is
used to get equivalence classes based on functional equivalence.
Since normal form reduction is undecidable in case of lambda
calculus, reduction steps are finitely bounded \cite{4}.

Formally a reaction between molecules $A$ and $B$ can be written
as a binary operation ($+_\Phi$) defined as

$$A +_\Phi B \rightarrow A + B + nf(((\Phi)A)B)$$

Where $+$ is used from the convention of writing chemical
equations to represent that the molecules are present in reactor.
$nf()$ uses some consistent reduction strategy to reduce the term
in finitely many steps to a normal form. This choice of finite
step normal form reduction actually results in equivalence classes
consisting of all the expressions which are functionally same
modulo finite execution steps. The choice of $\Phi$ gives
flexibility in the way molecules can react.\\\\
\textbf{Population Dynamics - Stochastic Molecular Collisions:}
Initially a large pool of random lambda terms of finite lengths is
generated. Only those terms, which are in normal form are
considered. In each iteration two molecules are chosen at random
and one is applied to the other (function composition) according
to $\Phi$, which is fixed at the beginning. Result is reduced to
its normal form in finite steps. Filtering conditions are applied,
for example, before collision takes place if the operator molecule
does not start with symbol $'\lambda'$ then it is discarded. These
filter conditions are basically meant to ensure consistency in
results as per the lambda calculus semantics and to give diversity
to the emerging organizational structures. Flow is maintained by
randomly selecting molecules and removing them from the reactor.

The relative quantitative dynamics of various molecules is
captured in terms of differential equations. Replicator equations
of Lotka-Volterra type \cite{3, 5} are used to describe the
relative concentration of self replicating molecules. \\\\
\textbf{Discussion:} In actual chemistry, especially in case of organic compounds with
 chains of carbon atoms and possible branching, chemical reactions
 substitute parts of one molecule with other molecule thus leading
 to structural rearrangement in the chemical composition of these
 molecules. This is the main motivation behind the choice of
 lambda terms in AlChemy, where the substitution is abstracted as function composition of lambda terms. Second motivation is that many chemical reactions can give rise to the same chemical compound, which is captured by normal form equivalence. The AlChemy is also a constructive chemistry like real chemistry.  Also the notion of equality leads to formation of network of molecules.

 With stochastic collision dynamics and choice of reaction type ($\Phi$), the AlChemy gives rise to some interesting forms of organizations, classified as \emph{level-0}, \emph{level-1}, and \emph{level-2} organizations. While \emph{level-0} organization consists of only self replicating molecules whose frequencies are modelled using replicator equations, \emph{level-1} organization has strong element of self-maintenance where any reaction between two molecules produces a new molecule inside the
 same population. \emph{level-2} organization is a coexistence of two interdependent \emph{level-1} organizations which support each other.

 Though AlChemy captures certain basic aspects of real chemical
 compounds and their reactions, it has its own limitations.
 Most important of those is related to the choice of lambda calculus.
 Even though lambda calculus is computationally universal and has a
 consistent reduction  strategy (i.e., order of reduction steps does not change the result),
 it has no serious bearing on its chemical counter part. Actual chemical reactions are not only much more complex,
 they might not follow computationally consistent mechanisms like total substitution.

 Thus the first limitation is the lack of
 selective substitution, which means, in case of actual
 chemical reactions, new compounds are formed (with substitution)
 based on the relative strengths of chemical bonds in reactants
 and relatively higher stability of the products. On the other hand in case of
 substitution in lambda terms no such conditions apply and instances
 of free variable are equally substituted everywhere.
 We propose alternate structure and reaction rules to overcome
 this limitation in the next section.

Second limitation is the poor abstraction of structural
    properties of chemical compounds. The only kind of compounds
    which might be resembling the lambda terms structurally are
    those which have long carbon chains with possible branching.
    Double helix structure of DNA
    with complementarity is difficult to capture using
    lambda terms. Other geometrical properties like chirality
    \footnote{Many important molecules required for life exist in two forms.
    These two forms are non-superimposable mirror images of each other,
    like the left and the right hand. This property is called
    chirality.}, which is so common in living
    forms \footnote{Nearly for all biological polymers to function
    their component monomers must have the same handedness.
    All amino acids in proteins are 'left-handed',
    while all sugars in DNA and RNA, and in the metabolic
    pathways, are 'right-handed'},
    as well cannot be captured using lambda terms. The significance of
    this lack of structural abstraction of geometrical properties is
    not very clear.

    Since chemical reactions are driven by thermodynamic
    constraints like rate of collision, pressure etc, and the properties of colliding
     molecules, they are usually symmetrical in nature. Thus the result of
     collision of the molecules A and B
     is same as that of B and A since there is no order
     on A and B.
      On the other hand, that is not the case with function composition,
    which is in general asymmetric in its definition.
    In our view, this presence of asymmetry
    in lambda chemistry might detach it from the real chemistry significantly.

    Functional Equivalence - the kind of functional
    equivalence defined in case of lambda chemistry does not
    capture the equivalence which life-like forms
    demonstrate. In case of living structures, it is the
    interaction which objects have with external environment or other
    objects that plays important role. This element of interaction is
    not captured well. One idea is to consider $\pi$ - calculus like
    formalism \cite{Parrow01} which has $bisimulation$ kind of
    equivalence which can be used to capture the equivalence in the objects based
    upon how they  can interact with other objects.

    Lack of information abstraction - this is true in
    general for almost all of the proposed ACs. And that is
    one of the focus of this proposal to understand the role
     information plays in the emergence of life-like phenomena in ACs.

\subsubsection{The Chemical Abstract Machine}~\label{cham}

The Chemical Abstract Machine (CHAM) was proposed in \cite{cham96}
as an abstract formalism for concurrent computation using closely
a metaphor of chemical reactions.

There are two description levels. On the upper level, CHAM
abstractly defines a syntactic framework and a simple set of
structural behavior laws. An actual machine is defined by adding a
specific syntax for molecule and a set of transformation rules
that specify how to produce new molecules from old ones. \\\\
\textbf{Molecules} are \emph{terms of some algebra}. A general
membrane construct transforms a solution into a single molecule,
and an associated general \emph{airlock} construct makes the
membrane somewhat porous to permit communication between an
encapsulated solution and its environment. The \emph{generic
reactions laws} specify how reactions defined by specific
transformation rules can take place and how membranes and airlocks
behave. A specific machine is defined by giving the algebra of
these terms and the rules. Not all molecules directly exhibit
interaction capabilities. Those which do are called \emph{ions}.
The interactive capability of an ion is generally determined only
by a part of it that is called its \emph{valence}.\emph{ The
reaction rules} are used to build new molecules from the ions. The
non-ion molecules can be heated as per the \emph{heating rules} to
break them into simpler sub-molecules. Conversely, a set of
molecules can cool down to a complex molecules using reverse
\emph{cooling rules}. The presence of membrane type structure
gives universal computational power to the model. \emph{Dynamics}
of CHAM goes like this - on each iteration a CHAM may perform an
arbitrary number of transformations in parallel, provided that no
molecule is used more than once to match the left side of a
reaction law. A CHAM is non-deterministic if more than one
transformation rules may be applied to the population at a time.

Sujuki and Tanaka used CHAM to model chemical systems by defining
an ordered abstract rewriting system on multiset called chemical
ARMS \cite{arms}. \emph{Molecules} are the \emph{abstract
symbols}. The \emph{reaction rules} are \emph{multiset rewriting
rules}. The reactor is represented by a multiset of symbols with a
set of input strings. An optional order is imposed on the rules,
which specifies in which order the rules are processed. Different
rate constants are modelled by different frequencies of rule
application. \\\\
The qualitative \textbf{dynamics} of ARMS is investigated by
generating rewriting rules randomly.  This led them to derive a
formal criteria for the emergence of cycles \cite{armscycles} in
terms of an order parameter, which is roughly the relation of the
number of heating rules to the number of cooling rules
\cite{armsorder}. For small and large values of this order
parameter, the dynamics remains simple, i.e., the rewriting system
terminates and no cycles appear. For intermediate values, cycles
emerge.\\\\
\textbf{Discussion:} Although CHAM was not defined as an AC, it is quite close to
actual cellular chemistry in some aspects. The presence of
membrane structure gives rise to important resemblance with
cellular reactions mediated by membranes. Another significant
property of CHAM model is that it is very general hence provides
flexibility in the way actual model is defined. Heating and
cooling laws closely capture what happens in case of actual
chemical reactions under the effect of temperature.

The main limitation of CHAM model is that the allowed abstract
terms of algebra are not adequate to capture the structural
properties of real chemical compounds, as discussed in case of
AlChemy.

Second limitation comes due to nature of rewriting rules, they are
actually grammar rules rather than being close to the chemical
reactions. Because of this problem with multiset rewriting, in
ARMS analysis is done by randomly generating these rewriting laws,
and it is not clear whether chemical reactions where molecules
actually interact and forge new bonds or break up can be fully
modelled this way.

\subsubsection{Artificial Chemistry on a Planar Graph}\label{pgraph}

This model of AC was proposed in \cite{planargraph}, where an AC
is embedded in a planar triangular graph. Molecules are placed on
the vertices of the undirected graph and interact with each other
only via the edges. The planar triangular graph can be manipulated
by adding and deleting nodes with a minimal local rearrangement of
the edges. The graph based approach provides handel for spatial
structures.\\\\
\textbf{Molecules and Reactions:} There is an (infinite) set of
potential molecules $S$ and a reaction mechanism which computes
the reaction product for two colliding molecules $x, y \in S$.
There may be an arbitrary number of products for each such
collision. Molecules are built from different types of substrate
of elements called atoms. Each type is associated with a different
function. The total number of atoms in the reactor is kept
constant during a run. Free atoms (not bounded in molecules) are
separately stored and form a global pool. \\\\
\textbf{Dynamics:} At every step they pick two neighboring molecules
$(x, y)$ and apply the first $x$ to the second $y$ creating a
(multi­)set of new molecules. These product molecules are randomly
inserted in the two faces next to the link between $x$ and $y$.
$x$ is replaced with first molecule after the reaction (the result
of the combinator reduction) and $y$ is finally deleted. Molecules
cannot change their positions in the graph.

In this system, it is observed that clusters of
molecules which do not interact with the neighboring molecules
arise. The clusters can be regarded as membranes when they divide
the graph into different regions. There also arises a cell
organization, that is, a subgraph that can maintain the membranes
by themselves.\\\\
\textbf{Discussion:} As noted in \cite{12}, the presence of spatial topology gives rise
to certain phenomena which is not possible to emerge easily in
cases where there is no spatial topology present in the model. For
example in the case of this planar graph based AC, an emergence of
membrane type structure is something which is frequently observed
in living systems. This phenomena does not emerge in open reactor
type of ACs with no spatial structures. Another important property
is the emergence of self organization in the form of maintaining
the membrane structures. Choice of "atom - symbols" as basic
molecular unit closely resembles real chemical composition of
molecules consisting of atoms.

On the other hand, the choice of planar graph based topology is
not something usually present in cellular structures neither it
can be a simplified spatial structure for initial chemical
environment responsible for emergence of life. Absence of
abstraction of geometrical or structural properties is yet another problem.

\subsection{More Discussion on Artificial Chemistries}

ACs are basically motivated by and developed to understand the
pre-biotic evolution or the problem of origin  of life, which is
still an open problem despite lots of advancements in molecular
biology \cite{6, 7, 8}. The problem of pre-biotic evolution
differs significantly from the post-biotic phenomena mainly
because of the appearance of genetic material. Once the first form
of life, a single cell or more primitive forms are available,
Darwinian theory of evolution based upon mutation and selection
\cite{9} or neutral theory of random drifts \cite{10}, etc can be
used to explain the emergence of higher and more complex forms of
life. Still the emergence of this genetic material which is so
fundamental for the proper functioning of even the simplest forms
of life is what makes the problem of pre-biotic evolution so
different.

Therefore the kind of problems mainly of focus in ACs and in this
proposal are the search for principles governing the emergence of
life-like forms from non life-like structures in AC systems. This
also involves proper level of abstraction from real chemistry
without loosing generality. 

In AC, we primarily consider the qualitative aspects of a problem,
before considering the quantitative relations between its
components. The quantitative aspect is usually analyzed using
reactor flow equations \cite{11}.  The stable structures generated
by artificial chemistries, the stable sets of molecules, are
usually referred to as \emph{organizations}. Understanding which
organization will appear is one such example to understand the
qualitative solution of an AC.

Some of the aspects very commonly studied in AC are - given an AC,
how to know a priori, which organizations are possible and which
are not possible? To know which organizations are probable and
which are improbable? To define an AC to generate a particular
organization? How stable are organizations? Can the complexity of
an organization be defined? If is possible to generate an AC which
moves from organization to organization in a never ending growth
of complexity? Quantitative questions can also be asked, for
example, given an AC, in a particular organization how many stable
(attractive) states are present inside it?

\cite{12} has detailed description of several interesting common
phenomena which are observed in different kinds of AC systems such
as reduction of diversity, formation of densely coupled stabled
networks, syntactic and semantic closure in these networks etc.

\section{P System based Artificial Graph Chemistry}

In the previous section we reviewed some examples of ACs with
discussion on their positive and negative aspects from the point
of view of pre-biotic evolution of life ranging from the level of
abstractions of essential molecular properties from real chemistry
to the nature of emerging organizational structures.

In this section we will propose a new AC to address the problem of
lack of structural abstraction in molecular structures and
reaction rules. Again as discussed in the previous section topological constraints play significant role in
emergence of certain phenomena in ACs, for example emergence of
membrane type structure, which limits the scope of molecular
interactions and promote local interactions. Nature of reaction
rule space as well has fundamental effect on the possible emerging
structures. We use extension of P system, explained next, to
capture the spatial membrane type topology for our AC. This should
enable us to understand better the role played by topological
constraints in emergence of life-like structures. \cite{arms} has
discussion on use of P-system to model pre biotic phenomena with
molecular structure and reaction rules as in case of ARMS (Section~\ref{cham}).


\subsection{P System}

G. Paun \cite{paunart} (ref \cite{paunbook}) introduced membrane
system as a model of parallel and distributed computation with
``membrane'' type structure. P system is a basic model of a
membrane systems with membranes arranged in a hierarchial
structure, as in a cell, and processing multisets of
symbol-objects.

\subsubsection*{Definition}

The main components of a membrane system are the \emph{membrane
structure}, \emph{multisets of objects}, and the \emph{evolution
rules}. A membrane structure describes the mutual relationship
between membranes, the relations of adjacency, of being in and
out.

Formally a membrane structure is defined as follows. Consider a
context free language $D$ defined over the alphabet $\{[, ]\}$ and
generated by following the grammar

$$S \rightarrow \epsilon | [S] | SS$$

Then language $MS$ over alphabet \{[, ]\} is defined as

$$MS = [D]$$

that is, MS consists of any string of correctly matching pair of
parentheses [, ] with a matching pair at the end.  $[_1 [_2 ]_2
[_3 ]_3 [_4 [_5 ]_5 [_6 ]_6 ]_4 ]_1$ is one such example where
brackets are numbered (with subscripts) to match the correct
pairing.

Let $\overline{MS}$ be the set of all equivalence classes of $MS$
with respect to reflexive and transitive closure of the relation
$\sim$ defined over $MS$ such that for $x, y \in MS$, $x \sim y$
if and only if $x$ and $y$ can be made same when two pairs of
parentheses which are not contained in each other are
interchanged, together with their contents. The elements of
$\overline{MS}$ are called \emph{membrane structures}.

Each matching pair of parentheses is called a \emph{membrane}.
There is a unique external membrane called \emph{skin}. For a
membrane structure $\mu$, any closed space delimited by a membrane
is called a \emph{region of $\mu$}. Alternately if we
 consider a membrane structure as a directed unordered rooted tree
then the depth of any membrane from the root (the skin) is called
its \emph{degree}. A membrane structure of degree $n$ contains $n$
regions, one associated with each membrane.

Consider $U$ as denumerable set of objects. Let $\mu$ be a
membrane structure of degree $n (\geq 1)$, with the membranes
labelled in a one ­to ­one manner, for instance, with the numbers
from $1$ to $n$. In this way, the regions of $\mu$ are also
identified by the numbers from $1$ to $n$. If a multiset $M_i: U
\rightarrow N$ is associated with each region $i$ of $\mu$, $1 < i
< n$, then we say that we have a \emph{super­cell}.

In other words a super-cell is defined as $(V, \mu, M_1, \ldots ,
M_n)$, where V is an alphabet; its elements are called objects.

Super cell extended to a \emph{P system} of degree $n$; $n \geq
1$, is a construct

$$\Pi = (V, \mu, M_1, \ldots , M_n, R_1, \ldots , R_n, ; i_0 ),$$

 where:

(i) $\mu$ is a membrane structure of degree $n$, with the
membranes and the regions labelled in a one ­to ­one manner with
elements in a given set $\Lambda$, which can be assumed to be just
$\{1, 2, \ldots, n\}$;

(ii) $M_i$, $1 \leq i \leq n$, are multisets over $V$ associated
with the regions $1, 2, \ldots , n$ of $\mu$;

(iv) $R_i$, $1 \leq i \leq n$, are finite sets of evolution rules
over $V$ . An evolution rule is of the form $u \rightarrow v$,
where $u$ is a string over $V$ and $v = v' | v'\delta$ , where
$v'$ is a string over

$$(V \times \{here, out\}) \cup (V \times \{in_j | 1 \leq j \leq n\}),$$

and $\delta$ is a special symbol not in $V$.

 (v) $i_0$ is a number between $1$ and $n$ which specifies the
 output membrane of $\Pi$.

The symbols $here,$ $out,$ $in_j$, $1 \leq j \leq n $ are called
\emph{target commands} or \emph{target indicators}. Length of $u$
in $u \rightarrow v$ is called the \emph{radius }of the rule.
These rules are explained in the next section.

\subsubsection*{Evolution of a P System}

The above defined P system evolves as follows. Consider a rule $u
\rightarrow v$ in a set $R_{i}$ . We look to the region associated
with the membrane $i$ . If the objects mentioned by $u$, with the
multiplicities specified by $u$, appear in $M_{i}$, then these
objects can evolve according to the rule. The rule can be used
only if it can use the objects in $u$. More precisely, we start to
examine the rules nondeterministically and assign objects to them.
A rule can be used only when there are enough copies of the
objects as specified by the definition of the rule, left after
application of other rules chosen non deterministically before
this rule. Therefore, all objects to which a rule can be applied
must be the subject of a rule application. All objects in $u$ are
consumed by using the rule $u \rightarrow v$, that is, the
multiset identified by $u$ is subtracted from $M_{i}$ .

The result of using the rule is determined by $v$. If an object
appears in $v$ in a pair $(a, here)$, then it will remain in the
same region $i$ . If an object appears in $v$ in a pair $(a,
out)$, then $a$ will exit the membrane $i$ and will become an
element of the region immediately outside it (thus, it will be
adjacent to the membrane $i$ from which it was expelled). In this
way, it is possible that an object leaves the super­cell itself:
if it goes outside the skin of the super­cell, then it never comes
back. If an object appears in a pair $(a, in_q )$, then $a$ will
be added to the multiset $M_q$ , providing that $a$ is adjacent to
the membrane $q$. If $(a, in_q )$ appears in $v$ and the membrane
$q$ is not one of the membranes delimiting ``from below'' the
region $i$ , then the application of the rule is not allowed.

If the symbol $\delta$ appears in $v$, then the membrane $i$ is
removed (said dissolved) and at the same time the set of rules
$R_i$ is removed. The multiset $M_i$ is added (in the sense of
multisets union) to the multiset associated with the region which
was immediately external to the membrane $i$. We do not allow the
dissolving of the skin, the outermost membrane because this means
that the super­cell is lost, we do no longer have a correct
configuration of the system.

A P system evolves as a whole unit with all possible applications
of rules at the same time. That means, all the operations are done
in parallel, for all possible applicable rules $u \rightarrow v$,
for all occurrences of multisets specified by $u$ in the region
associated with the rule and all regions are considered at the
same time.

 If there are rules in a super­cell system $\Pi$ with the radius at
least two, then the system is said to be \emph{cooperative}; in
the opposite case, it is called \emph{non­cooperative}. A system
is said to be \emph{catalytic} if there are certain objects $c_1,
\ldots, c_n$ specified in advance, called \emph{catalysts}, such
that the rules of the system are either of the form $a \rightarrow
v$, or of the form $c_i a \rightarrow c_i v$, where $a$ is a
non­catalysts object and $v$ contains no catalyst. (So, the only
cooperative rules involve catalysts, which are reproduced by the
rule application, and left in the same place. There are no rules
for the separate evolution of catalysts.

\emph{Biological analogy of Super Cell system.} The mode of
evolving of objects in a super­cell provided with evolution rules
as described above can be interpreted in the following - idealized
- biochemical way. We have an organism, delimited by a skin (the
skin membrane). Inside, there are free molecules, organized
hierarchically. The molecules float randomly in the ``cytoplasmic
liquid'' of each membrane. Under specific conditions, the
molecules evolve, alone or with the help of certain catalysts.
This is done in parallel, synchronously for all molecules. The new
molecules can remain in the same region where they have appeared,
or can pass through the membranes delimiting this space,
selectively. Some reactions not only modify molecules, but also
break membranes. When a membrane is broken, the molecules
previously placed inside it will remain free in the larger space
newly created, but the evolution rules of the former membrane are
lost. The assumption is that the reaction conditions from the
previous membrane are modified by the disparition of the membrane
and in the newly created space, only the rules specific to this
space can act.  When the external membrane is broken, then the
organism ceases to exist.

\subsection{Design of the Chemistry}

Having defined the basic concepts of P System, we will now discuss 
the basic components of the P system based artificial graph chemistry (AGC) 
by defning an extension of the P System and the molucules as labelled graph.

\subsubsection{A Probabilistic Extension of P System}

We extend a P system by associating probabilities with rules.
These probabilities can be interpreted as relative frequencies of
rule applications on the molecules over the course of evolution.
This happens in real chemistries as well, where in a large pool of
several chemical compounds (which was supposedly the situation in
pre-biotic world) the reactions which take place are largely
random and possibly only affected by the neighborhood of a
molecule to some extent. The advantage of associating
probabilities is to capture more clearly how relative frequencies
of application of certain rules affects the kind of emerging
structure in later states of evolution.
 The main role of this extension comes into picture when an AC on P
 system is evolved. Under this scheme molecules will be distributed
 in regions and during the long course of evolution of these regions
 and populations inside them,
 reaction rules are applied based upon the probabilities associated
 with them. This differs from the
 original static structure of rule space and resemble more like
 naturally occurring processes.

Another main modification on the basic definition of P system is
that to suit an AC set up, we modify the basic P system with
multisets of symbols and rewriting rules to a P system with
multisets of molecules with the reaction rules. The structure of
molecules with reaction semantics is presented next.

\subsubsection{Molecular Structure Representation and Reaction Semantics}

 We represent a molecule as undirected labelled graph and develop
reaction semantics to represent molecular reactions. The main motivation behind the selection of a graph comes from the
observation on the lack of the structural abstractions in ACs like
lambda chemistry, chemical abstract machine etc.

 Formally a \emph{molecule} is represented as a \emph{undirected labelled
graph} $G = (V, E)$ with mapping $w: E \rightarrow R$ associating
weights with edges $e \in E$, where $R$ is the set of real
numbers. $G$ can also be represented as weighted symmetric matrix.
Each node of the graph can be thought of as an ``atom'' and an
edge as ``chemical bond'' such that label/weight associated with
an edge determines the relative strength of the bond between these
atoms. These weights will be used to decide the possible
structural changes (substitution etc) during reactions.

Each \emph{reaction rule} is a \emph{mapping} from a subset of
molecules called \emph{reactants} to another subset of
\emph{molecules} called \emph{products}. In other words a reaction
can be thought of as a $n$-ary ($n \geq 2$) \emph{graph
transformation operator}. Reactions are constrained by
\emph{guards}, which are used to capture the thermodynamic
conditions controlling the real chemical reactions.

We can consider two cases. In the first case the set of possible
reaction rules is fixed and determined priori. Only the guards are
evaluated later. Secondly we can have some generic laws
controlling the nature of possible reactions, which actually
happens in the case of real world, where concentration,
temperature, pressure, velocity and other kinetic properties
determine what will be the result of the reactions. We need to
understand how different choices of these reaction rules affect
the information processing, computational structure and what kind
of organizational structures emerge during the course of
evolution.

Initially we choose no specific spatial structure on distribution
of molecules. Thus in essence each molecule can potentially react
with any other molecule. The dynamics is controlled
stochastically, that is, we randomly select the rules and see if
they can be carried out, if yes, we select random concentration of
the molecules for reaction. Though we have no explicit spatial
structure but we want that not all the reactants get over with
single rule application, so we choose this strategy of choice of
random concentrations. We replace the reactant molecules with the
products in the reactor. Like CHAM we consider the heating and the
cooling rules, which result in braking up of the molecule-graph
into smaller components, and joining of smaller graphs
respectively.

A spatial structure can be imposed on the molecules using our
extended probabilistic P-system, where sets of molecules will be
encapsulated inside membranes for local reactions, with possible
migration of molecules across membranes.

\section{Final Discussion}

Objective of this section is to summarize main goals of the
proposal and discuss the broader picture where these goals may fit
in an AC research.

To summarize, we defined a probabilistic P system based AGC with the aim of understanding the
principles leading to the evolution of life-like structures in an AC set up. We need to explore it further by carrying out detailed
experiments with varying parameters such as definitions of reaction rules, presence and nature of topological constraints,
distribution of molecules, and population dynamics. 

ACs are basically designed to complement the main stream AL
research. This is primarily because major AL studies presume the
prior existence of basic structure of life-like entities and
develop over them. This leaves the question of origin of these
basic structures open and that is where ACs come into picture.

Because the main theme of AL research is to discover the possible
biological principles which might be working independent of
physical laws, AL studies mainly draw motivation from real-life
biological phenomena. Theory of evolution based on random
mutations and fitness based natural selection is one such source
of motivation in many AL studies \cite{1}. Similarly ACs also draw
motivation from real chemistry. \emph{The main conceptual
motivation ACs borrow from real chemistries is not the actual
chemical structures or reactions but the abstract concept that
life originated as a result of complex dynamical interplay between the rule
space consisting of reaction rules or semantics and the object space consisting
of the molecules which react.} This is what is the prime source of
differences in various ACs in their definition and structures,
since there is no such generic framework which can used by ACs to
define the basic structure of molecules or reactions. Most often
what is clear is only the end results, that is, an AC set up is
expected to lead to the emergence of certain basic characteristics
of life, e.g., self-replication, metabolism etc.

This is where this proposal is expected to contribute most by
explaining issues like what are the fundamental ingredients of an
AC set up which will lead to interesting emergent organizations?
Can these ingredients be related to information and/or
computation? Is an AC able to create information? Does
``information processing'' emerge by way of evolution? Can
molecules be interpreted as computational units which evolve in
computational power during the course of evolution of an AC? What
kind of complex computational structures emerge and how? Has
communication any role to play in origin of organizational
structures? What kind of limits one can put on the basis of the
basic structure defined in an AC  on the possible level and type
emerging organizations? Thus as conclusion we expect that the proposal will contribute
 significantly in the research on ACs.

\end{document}